\documentclass[letterpaper, 10 pt, conference]{ieeeconf}  
\usepackage[utf8]{inputenc}

\IEEEoverridecommandlockouts                              

\usepackage{graphicx} 
\usepackage{times} 
\usepackage{amsmath} 
\usepackage{amssymb}  
\usepackage{dsfont}
\usepackage{todonotes}
\usepackage{booktabs}
\usepackage{siunitx}
\usepackage{subcaption}
\usepackage{hyperref}

\usepackage{algorithm, algorithmicx}
\usepackage[noend]{algpseudocode}

\newcommand{\ra}[1]{\renewcommand{\arraystretch}{#1}}

\newcommand{\failInd}[2]{\mathds{1}_{r(#1, \varphi) < #2}}

\title{\LARGE \bf Testing Rare Downstream Safety Violations via Upstream Adaptive Sampling of Perception Error Models}

\author{Craig Innes and Subramanian Ramamoorthy$^{1,2}$
\thanks{$^{1}$Craig Innes ({\tt\small craig.innes@ed.ac.uk}) and Subramanian Ramamoorthy ({\tt\small s.ramamoorthy@ed.ac.uk}) are with the School of Informatics,
        University of Edinburgh, Edinburgh, United Kingdom. 
        }%
\thanks{$^{2}$  Work supported by a grant from the UKRI Strategic Priorities Fund to the UKRI Research Node on Trustworthy Autonomous Systems Governance and Regulation (EP/V026607/1, 2020-2024). For the purpose of open access, the author(s) has applied a Creative Commons Attribution (CC BY) license to any Accepted Manuscript version arising}}

\begin{document}

\maketitle
\thispagestyle{empty}
\pagestyle{empty}

\begin{abstract}
Testing black-box perceptual-control systems in simulation faces two difficulties. Firstly, perceptual inputs in simulation lack the fidelity of real-world sensor inputs. Secondly, for a reasonably accurate perception system, encountering a rare failure trajectory may require running infeasibly many simulations. This paper combines perception error models---surrogates for a sensor-based detection system---with state-dependent adaptive importance sampling. This allows us to efficiently assess the rare failure probabilities for real-world perceptual control systems within simulation. Our experiments with an autonomous braking system equipped with an RGB obstacle-detector show that our method can calculate accurate failure probabilities with an inexpensive number of simulations. Further, we show how choice of safety metric can influence the process of learning proposal distributions capable of reliably sampling high-probability failures.
\end{abstract}

\section{INTRODUCTION}

Perceptual-control systems, such as in automated vehicles (\textsc{av}), must be rigorously evaluated before deployment. Safety specifications can be defined formally in Signal Temporal Logic (\textsc{stl}) \cite{hekmatnejad2019encoding, arechiga2019specifying}, but the black-box nature and operating scenario complexity of such systems often preclude analytical guarantees. Instead, we seek assurances via testing.

\begin{figure}
    \centering
    \includegraphics[width=0.7\linewidth]{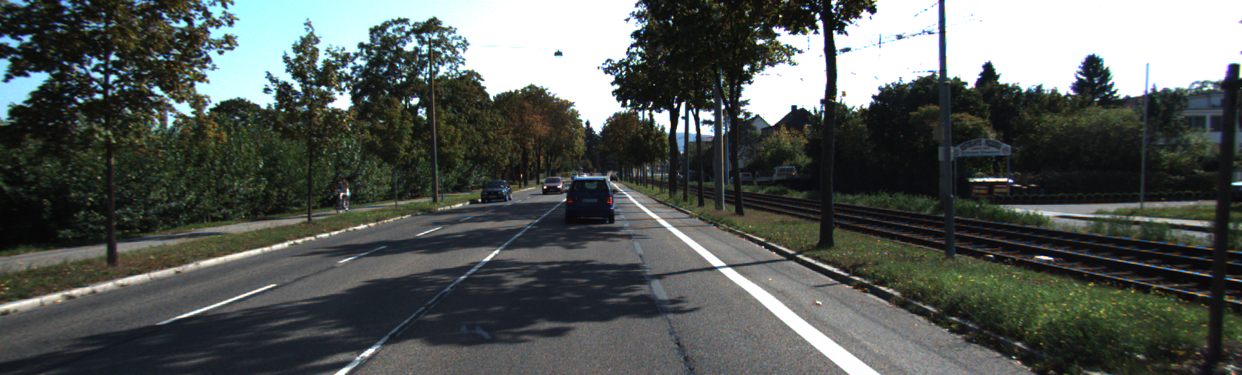}
    \includegraphics[width=0.7\linewidth]{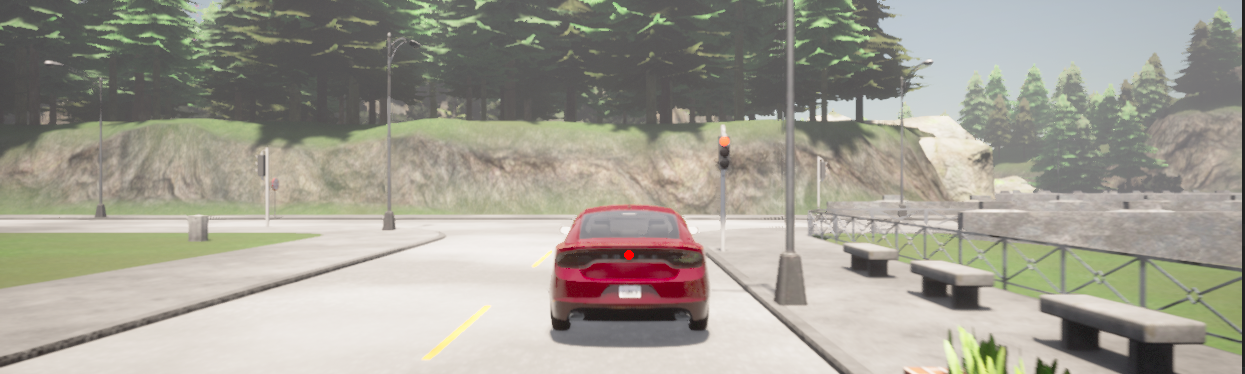}
    \caption{The difference in fidelity between real traffic images (top) and  simulated images rendered in CARLA (bottom).}
    \label{fig:photorealism}
\end{figure}

While one can aim at the ideal of conducting exhaustive real world tests, this is infeasible \cite{kalra2016driving}. In addition to the cost, these tests represent danger to the public. Alternatively, we can simulate our system, but simulation poses two core problems: First, while modern simulators can replicate real-world physics with high accuracy, perceptual inputs from a simulated sensor are far from their real-world counterparts. Figure \ref{fig:photorealism} shows the photorealism gap between a real and simulated RGB-camera image. In a black-box system with a mixture of perception and control components, this means that, if we train the upstream perception components on simulated inputs, we lack assurances that simulated downstream control failures will correspond to failures experienced by our system when using real-world sensor inputs.


However, a key insight is that, to trust our simulation, photorealism may be unnecessary. It is sufficient that the simulated perception system is accurate in the same situations as the real one. \emph{Perception Error Models} (\textsc{pem}s) \cite{sadeghi2021step} are surrogate models of a perception system which, instead of taking sensor inputs, take salient characteristics of those inputs, outputting the expected accuracy of the target system.

A second problem is, for any modestly accurate perception system, safety violations can be extremely rare -- especially if a system must fail many times in sequence to cause danger. Consider an automated braking system with an image-based obstacle detector. If the detector achieves $99\%$ accuracy, and needs to miss an obstacle for one second to cause a crash, then at 20 images-per-second, failure probability is $(1.0 - 0.99)^{20}$. This means we would expect to run $\num{1e40}$ simulations before observing \emph{a single crash}.

Instead of direct sampling, \emph{importance sampling} draws from a proposal distribution where failures are likely, then re-weights samples according to their likelihood under the target. The challenge is finding proposals where failures are likely, while producing trajectories common under the original distribution. Existing importance sampling approaches define proposal distributions over full trajectories, often failing to produce proposals capable of generating extremely rare events caused by long sequences of perception failures.

This paper proposes a method for estimating black-box perceptual-control system failures against \textsc{stl} specifications in simulation. To achieve this, we first train a \textsc{pem} to mimic a perception system trained on real-world images. Then, we exploit this \textsc{pem} to drive learning of a low-variance sampler using \emph{state-dependent} adaptive importance sampling.

We apply our method to an automated braking scenario in the \textsc{carla} simulator. The perceptual-control system under test uses a \textsc{yolo} obstacle detector \cite{redmon2018yolov3} trained on images from the \textsc{kitti} dataset \cite{geiger2013vision}. Our experiments show that, with a representative set of training images and sufficiently descriptive salient variables, our method accurately estimates how upstream perception accuracy affects downstream safety violations. Since multiple robustness metric choices exist for \textsc{stl}, our experiments also demonstrate how different metrics affect the learning process of adaptive sampling.

\section{ESTIMATING THE PROBABILITY OF FAILURE}

We first formally describe failure probability estimation for our perceptual-control system: Our environment contains states $s$ in $d$-dimensional space $\mathcal{S} = \left\{ (x_1, \dots, x_d) \mid x_{1\dots d} \in X_{1\dots d}\right\}$. Our perceptual-control system $\pi: \mathcal{S} \times \mathcal{A} \rightarrow [0, 1]$ will, given state $s$, output action $a \in \mathcal{A}$ with probability $\pi(a | s)$. In the \textsc{av} domain, variables $X_1, \dots, X_d$ may include vehicle position/velocity, or pixel values from a camera; action space $\mathcal{A}$ may consist of whether an obstacle detection system detects a vehicle in frame.

The probability of transitioning to state $s'$ from $s$ is $p(s', a \mid s) = \pi(a \mid s) p(s' \mid s, a)$. This paper makes the simplifying assumption (as in \cite{corso2020scalable}) that the only source of stochasticity is perception system outputs. E.g., once the perception system decides an obstacle is ahead, the decision to brake (and effect of braking on velocity) is deterministic:

\begin{equation}
p(s' , a \mid s) = \pi(a \mid s)
\end{equation}

Starting from $s_0$ we define a $T$-step \emph{simulation rollout} as $\tau = \left[ s_0, a_1, s_1, a_2 \dots s_{T - 1} \right]$, where the rollout probability is:

\begin{equation}
    p(\tau) = \prod_{t=1}^{T-1} \pi(a_t \mid s_{t-1})
    \label{eqn:traj-prob}
\end{equation}

We define a \emph{safety metric} $r(\varphi, \tau) \in \mathbb{R}$, where $\varphi$ is a specification, and $r(\varphi, \tau) < 0$, denotes violation. Our goal is to compute the probability our system falls below some safety threshold $\gamma$: 

\begin{equation}
    \mathbb{E}_{\pi}\left[ \mathds{1}\left\{r(\mathbf{\tau}, \varphi) \leq \gamma \right\} \right] = \int_{\tau} \mathds{1}\left\{ r(\mathbf{\tau}, \varphi) \leq \gamma \right\}p(\tau)
    \label{eqn:expected-fail}
\end{equation}

Subsequent sections describe the components needed to estimate (\ref{eqn:expected-fail}) --- specifying $\varphi$ and $r(\tau, \varphi)$, constructing a \textsc{pem}, and learning a proposal distribution.

\subsection{Specifying Safety Requirements with STL}
\label{sec:stl}

Signal Temporal Logic (\textsc{stl}) is a language for expressing properties of timed, continuous signals \cite{donze2010robust}. Its grammar is:

\begin{equation}
    \begin{aligned}
        &\varphi := \top \mid \eta \mid \neg \varphi \mid \varphi_1 \wedge \varphi_2 \mid \square_{I} \varphi \mid \diamondsuit_{I} \varphi \mid \varphi_1 \mathcal{U}_{I} \varphi_2 \\
    \end{aligned}
\label{eqn:stl-syntax}
\end{equation}
where $\eta$ is any predicate $\rho(s) - b \leq 0$ (with $b \in \mathbb{R}$ and $\rho: \mathcal{S} \rightarrow \mathbb{R}$). Symbol $\square_{I} \varphi$ means  $\varphi$ is \emph{always} true at every time-step within interval $I$. Symbol $\diamond_{I} \varphi$  means $\varphi$ must \emph{eventually} be true at some time-step in $I$, and $\varphi_{1} \mathcal{U}_{I} \varphi_{2}$ means $\varphi_1$ must remain true within $I$ \emph{until} $\varphi_{2}$ becomes true. For example:

\begin{equation}
    \square_{[0, T]} \left( \lVert C_{ego}.pos - C_{other}.pos \rVert \geq 2.0 \right)
    \label{eqn:stl-vehicle-distance}
\end{equation}
means \emph{``From 0 to T, the distance between the ego car and other car must never drop below 2 metres''}. \textsc{stl} formulae are paired with a \emph{robustness metric} $r(\tau, \varphi) \in \mathbb{R}$, specifying how strongly $\tau$ satisfied $\varphi$ ($r(\tau, \varphi) > 0$) or violated $\varphi$ ($r(\tau, \varphi) < 0$). Classical robustness focuses on \emph{spatial robustness}, applying hard minimums across conjunctions \cite{donze2010robust}. For instance, equation (\ref{eqn:classical-always}) evaluates the \emph{always} operator:

\begin{equation}
    \label{eqn:classical-always}
    r(\tau, \square_{[0, T]}\varphi) = \min \left ( r(s_0, \varphi), \dots, r(s_T, \varphi) \right)
\end{equation}
Applied to (\ref{eqn:stl-vehicle-distance}), this corresponds to the single state in $\tau$ with the smallest distance between vehicles.

Alternative robustness metrics exist, each offering trade-offs in time/space robustness, soundness, and smoothness: \textsc{agm}-robustness \cite{mehdipour2019arithmetic} replaces the min/max functions in \emph{always}/\emph{eventually} with geometric/arithmetic means; Smooth Cumulative Robustness \cite{haghighi2019control}, combines smooth-approximations to min/max with cumulative summations.

\subsection{Perception Error Models}
\label{sec:pems}

In equation (\ref{eqn:traj-prob}), $\tau$ is determined by our perceptual model $\pi(\mathcal{A} \mid s)$. For an obstacle detector, this could be a function $f$ over RGB images, outputting $detected$ or $\neg detected$:

\begin{equation}
    \pi(\mathit{detected} \mid s) = f_{\theta}(s)
    \label{eqn:detection-f}
\end{equation}
Here $\theta$ could represent e.g., the weights of a Convolutional Neural Network trained on real traffic images. When testing in simulation however, we cannot use $f_{\theta}$ directly, since our simulator does not produce real sensor inputs. We could train an alternative $f$ based on rendered images, but as discussed in the introduction, such images correspond poorly to the real world. The added cost of including a rendering and image-processing step in the simulation pipeline also increases simulation costs by an order of magnitude. 

Here is the key insight of \textsc{pem}s: For simulated risk assessment to be useful, it is sufficient that, for the same state $s$, our surrogate $\hat{f}$ and function $f$ share the same error:

\begin{equation}
    \forall s \in \mathcal{S}, f_{\theta}(s) = \hat{f}_{\hat{\theta}}(g(s))
\end{equation}
where $g$ is a projection of $\mathcal{S}$ (which could contain RGB image pixels, or other sensor data), down to a set of \emph{salient variables} --- a subset of perceptual features necessary to predict system performance. This may include obstacle distance to camera, vehicle type, or level of occlusion.

We train $\hat{f}$ using $m$ training samples $D$, consisting of predictions from our perception system $f$ paired with the corresponding salient variables:

\begin{equation}
    D = \{\langle f(s_0), g(s_0) \rangle, \dots \langle f(s_m), g(s_m) \rangle \}
\end{equation}

We then find the $\hat{\theta}$ which minimizes the binary cross entropy (\textsc{bce}) between predictions $f_{\theta}(s)$ and $\hat{f}_{\hat{\theta}}(g(s))$:

\begin{equation}
    \sum_{i=0}^{m} f(s_i) \log\left[\hat{f}_{\hat{\theta}}( g(s_i))\right] + (1 - f(s_i)) \log \left[1 - \hat{f}_{\hat{\theta}}( g(s_i)) \right]
    \label{eqn:pem-bce}
\end{equation}

Once $\hat{f}_{\hat{\theta}}$ is learned, we can use it as a drop-in surrogate for $f$ in our simulator by projecting each $s$ down to its salient variables as they arrive online.

\subsection{Adaptive Importance Sampling}
\label{sec:ais}

Given $N$ simulations, we can estimate (\ref{eqn:expected-fail}) via monte-carlo approximation:

\begin{equation}
    \hat{\mu}_{\textsc{mc}} = \frac{1}{N} \sum_{i=0}^{N} \mathds{1}_{\left\{ r(\mathbf{\tau_{i}}, \varphi) \leq \gamma \right\}}
    \label{eqn:monte-carlo}
\end{equation}
For rare-events, this is highly inefficient. For $N$ samples, the relative error \cite{kroese2013cross} is approximately $\sqrt{(N \mu)^{-1}}$. This means to achieve a $1\%$ relative error with event probability $\mu = 10^{-6}$, we need to run $10^{10}$ simulations. A more efficient approach is to use \emph{importance sampling} to draw from a proposal $q$ where the rare-event is more likely, then re-weight each sample with respect to target distribution $p$:

\begin{equation}
    \hat{\mu}_{\textsc{is}} = \frac{1}{N} \sum_{i=0}^{N} \mathds{1}_{\left\{ r(\mathbf{\tau_{i}}, \varphi) \leq \gamma \right\}} \frac{p(\tau_i)}{q(\tau_i)}
\end{equation}

Rather than defining distributions over $\tau$, our modified importance sampler defines proposals over \emph{states} $s \in \mathcal{S}$:

\begin{equation}
    \frac{1}{N} \sum_{i=0}^{N} \mathds{1}_{\left\{ r(\mathbf{\tau_{i}}, \varphi) \leq \gamma \right\}} \prod_{t=0}^{T} \frac{\hat{f}_{\hat{\theta}}(g(s_{i, t}))}{q_{\phi}(h(s_{i, t}))}
    \label{eqn:is-factored}
\end{equation}
We use $s_{i,t}$ to denote state at time $t$ of the i-th simulation. Following (\ref{eqn:traj-prob}) and (\ref{eqn:detection-f}), we factor $p(\tau)$ as a product of state-dependent outputs from \textsc{pem} $\hat{f}$. Likewise, $q(\tau)$ is factored as product of probabilities $q_{\phi}(h(s))$. Note, $h(s)$ does not need to be the same as $g(s)$. $h(s)$ might project $s$ over a small subset of features we suspect influence safety e.g., vehicle acceleration or velocity.

We want a proposal where failure events are common, and trajectories have high likelihood under the target distribution. The perfect proposal $q^*$ has the form:

\begin{equation}
    q^{*}(\tau) = \left(\frac{p(\tau)}{\mu} \right) \mathds{1}_{r(\tau, \varphi) < \gamma}
    \label{eqn:perfect_estimator}
\end{equation}
Of course, we do not know $\mu$, but given example trajectories $[ \tau_0, \dots \tau_N ]$, we can learn good proposal parameters $\phi$ by minimizing the KL-divergence between the data and (\ref{eqn:perfect_estimator}): 

\begin{equation}
    -\sum_{i=0}^{N} w_i \left(\sum_{t=0}^{T} \log q_{\phi}(h(s_{i, t})) \right)\failInd{\tau_i}{\gamma}
    \label{eqn:importance-kl}
\end{equation}
with weights $w_i$ defined as: 

\begin{equation}
    w_i = \prod_{t=0}^{T} \frac{\hat{f}_{\hat{\theta}}(g(s_{i, t}))}{q_{\phi}(h(s_{i, t}))}
\end{equation}
However, we have a problem: If no trajectory in $\tau_{0}, \dots, \tau_{N}$ passes threshold $\failInd{\tau}{\gamma}$, this halts optimization. To address this, we apply techniques from the cross-entropy method (\textsc{cem}) literature \cite{kroese2013cross} --- Instead of filtering all simulations by threshold $\gamma$, we iteratively sample multiple batches of simulations. At each stage $\kappa \in [0, \dots K]$, we run $N_{\kappa}$ simulations, then sort $\{ \tau_1, \dots, \tau_{N_{\kappa}}\}$ by $r(\tau, \varphi)$. In place of $\gamma$, we set an intermediate threshold $\gamma_{\kappa} = \max(\gamma, r(\tau_{\sigma N_{\kappa}}))$, where $\sigma$ is a desired quantile within range $0.95 \leq \sigma < 1.0$. This ensures some simulations meet the threshold at every stage, driving incremental proposal learning via (\ref{eqn:importance-kl}).

Because we are finding a proposal via numerical optimization of a product of state dependent distributions (rather than e.g., a single exponential distribution admitting a closed-form solution \cite{kim2016improving, zhao2016accelerated, o2018scalable}), and because low safety thresholds are characterized by long chains of consecutive failures, we have another problem: The difference between the highest and lowest trajectory weights $w$ may become enormous. This difference can lead to numerical instability, with less likely trajectories having effectively zero weight during learning. If those less likely trajectories are necessary to achieve lower safety thresholds, this can result in the optimization stalling at an intermediate threshold $\gamma_{\kappa}$.

Defensive Importance Sampling techniques can be used to limit the ratio between the target and proposal likelihoods for a single state \cite{bugallo2017adaptive}, but such techniques place no limit on weights $w_i$ resulting from the \emph{product} of likelihood sequences. Instead, our approach is to apply a smoothing parameter $\alpha \in [0, 1]$ to importance weights:

\begin{equation}
    -\sum_{i=0}^{N} w_i^{\alpha} \left(\sum_{t=0}^{T} \log q_{\phi}(h(s_{i, t})) \right)\failInd{\tau_i}{\gamma}
    \label{eqn:smooth-is}
\end{equation}
This lessens the gap between the highest and lowest weights, allowing our adaptive learning process to progress.

\begin{algorithm}
        \caption{Simulated Risk Estimation with \textsc{pem}s}
        \label{alg:full-system}
        \begin{algorithmic}[1]
        \Function{adaptive-est}{$D, \varphi,N_{\kappa}, N_{e}, \gamma$}
            \State $\hat{\theta} \gets $ Train \textsc{pem} using $D$ with (\ref{eqn:pem-bce})
            \State $\phi \gets$ Initialize proposal distribution
            \For{$\kappa = 1$ to $K$}
                \State $\{ \tau_{0} \dots \tau_{N_\kappa} \} \gets$ Sample $N_{\kappa}$ rollouts with $q_{\phi}$
                \State Sort $\{ \tau_{0} \dots \tau_{N_\kappa} \}$ by $r(\tau, \varphi)$
                \State $\gamma_{\kappa} \gets max\left(\gamma, r\left(\varphi, \tau_{\lfloor\sigma N_{\kappa}\rfloor}\right)\right)$
                \State $\phi \gets$ Min (\ref{eqn:smooth-is}) with $\gamma_{\kappa}, q_{\phi}, \hat{f}_{\hat{\theta}}, \{\tau_{0} \dots \tau_{N_{\kappa}}\}$
            \EndFor
            \State $\{ \tau_{0} \dots \tau_{N_e} \} \gets$ Sample $N_{e}$ rollouts with $q_{\phi}$
            \State \Return $\hat{\mu_{\textsc{is}}}$ using $\{\tau_{0} \dots \tau_{N_e} \}$ with (\ref{eqn:is-factored})
        \EndFunction
        \end{algorithmic}
\end{algorithm}

Algorithm (\ref{alg:full-system}) Combines our \textsc{stl} robustness components, \textsc{pem}, and adaptive sampling to give our full method. After training our \textsc{pem} to replicate real-world perceptual system behaviour as outlined in section (\ref{sec:pems}), we perform $K$ stages of black-box simulation. At each stage $\kappa$, we conduct $N_{\kappa}$ simulations, set threshold $\gamma_{\kappa}$ based on the upper quantile of rollouts, then use these rollouts to optimize parameters $\phi$ for the next iteration. After the $K$ iterations, we perform a final $N_{e}$ simulations to estimate failure probability $\mu$.

\section{EXPERIMENTS}

Our experiments apply algorithm (\ref{alg:full-system}) to an automated braking problem in the CARLA vehicle simulator \cite{dosovitskiy2017carla}. We show our method can estimate the probability of a rare collision event efficiently, yielding a greater number of high-probability failure events (compared with classical monte-carlo sampling, or a naive custom proposal). Further, we show how the choice of \textsc{stl} robustness metric can affect learning of the proposal distribution.

\begin{figure}
    \centering
    \includegraphics[width=0.5\linewidth]{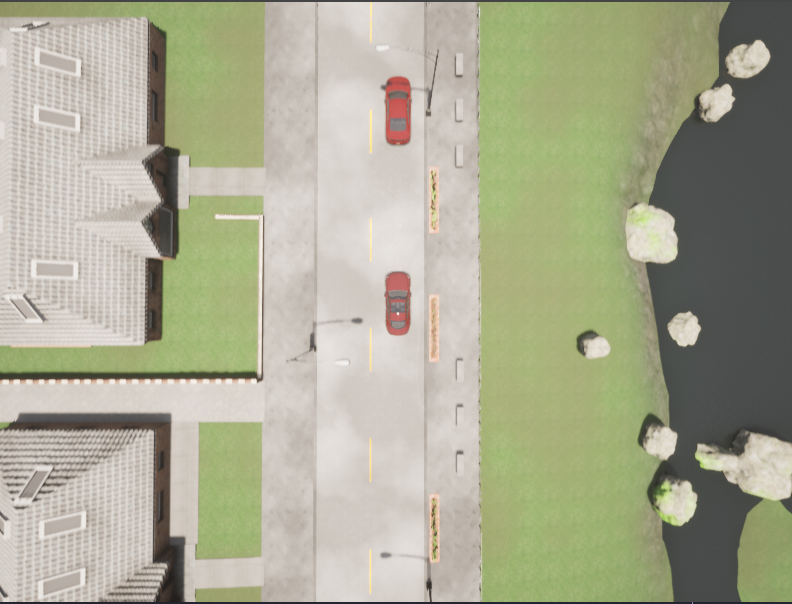}
    \caption{Automated braking scenario --- When the front car abruptly stops, our vehicle must brake to avoid crashing.}
    \label{fig:CARLA-scenario}
\end{figure}

Figure (\ref{fig:CARLA-scenario}) illustrates our scenario: Our car follows another for $T=100$ time steps. At the end of the road, the front vehicle brakes, and our car must brake to avoid crashing. We use (\ref{eqn:stl-vehicle-distance}) as our safety condition --- If the distance between vehicles is ever less than 2 metres we consider it a ``crash''\footnote{Note, the true distance is less than 2 metres, as we measure from the windscreen camera to the rear of the other car \cite{geiger2013vision}}.

Our vehicle's perceptual-control system comprises two main components: a physical controller and a perception system. The physical controller is a simplification of the \texttt{BasicAgent} from the base CARLA library\footnote{\url{https://carla.readthedocs.io}}: A global planner sets road waypoints for the car to follow; A local planner applies \textsc{pid} control to reach max speed if the road is obstacle-free, and applies emergency brakes if there is an obstacle within 10 metres of the vehicle\footnote{Code at \url{https://github.com/craigiedon/CarlaStuff}}. The perception system is a neural-network-based obstacle detector trained on RGB images of real traffic from the \textsc{kitti} dataset \cite{geiger2013vision}. We first describe how we train a \textsc{pem} from this detector, then outline its use within CARLA.

\subsection{Learning a \textsc{pem} of the \textsc{kitti} dataset}

\textsc{kitti} is an object detection dataset of $1242 \times 375$ traffic images, with labels denoting vehicle type and 2D bounding boxes. It contains 7481 training and 7518 test images.

\begin{figure}
    \centering
    \includegraphics[width=0.7\linewidth]{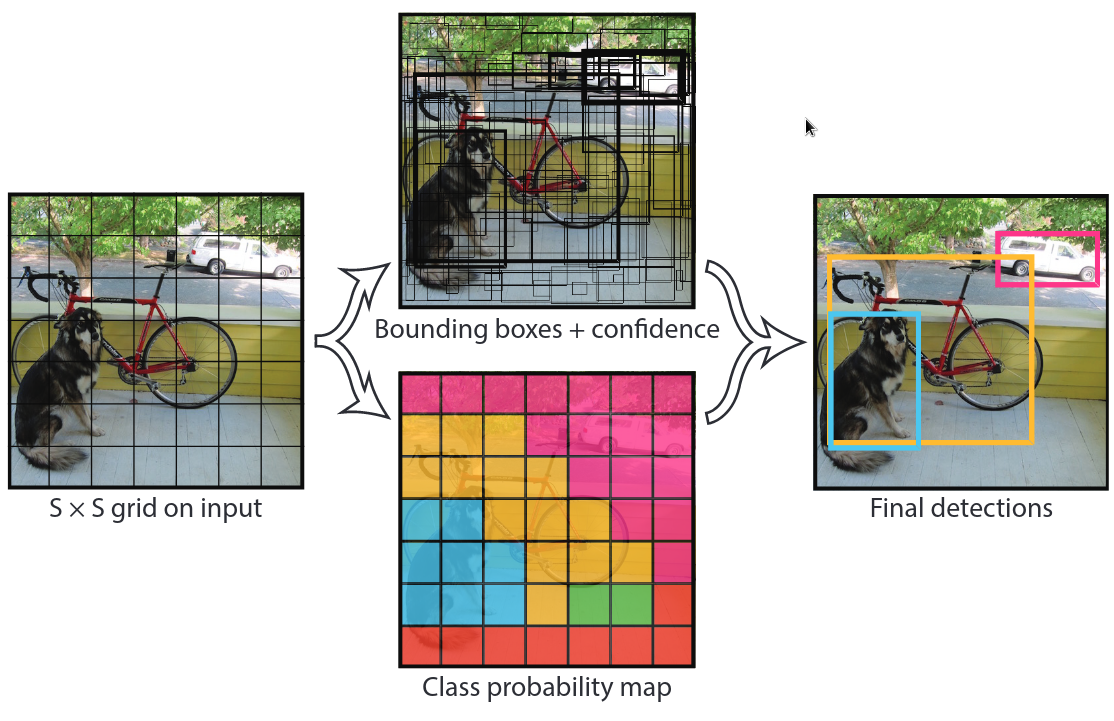}
    \caption{(Diagram from \cite{redmon2018yolov3}) The \textsc{yolo} detector. Images are divided into a grid. For each grid cell, it predicts B bounding boxes, box confidences, and class probabilities.}
    \label{fig:yolo-diag}
\end{figure}

The obstacle detection system uses \textsc{yolo-v3} \cite{redmon2018yolov3}, trained on \textsc{kitti}. Figure \ref{fig:yolo-diag} sketches \textsc{yolo}'s operation. Images are divided into regions, bounding boxes produced, and detection probabilities $f(s)$ predicted in one pass through the network.

\begin{figure}
    \centering
    \includegraphics[width=0.85\linewidth]{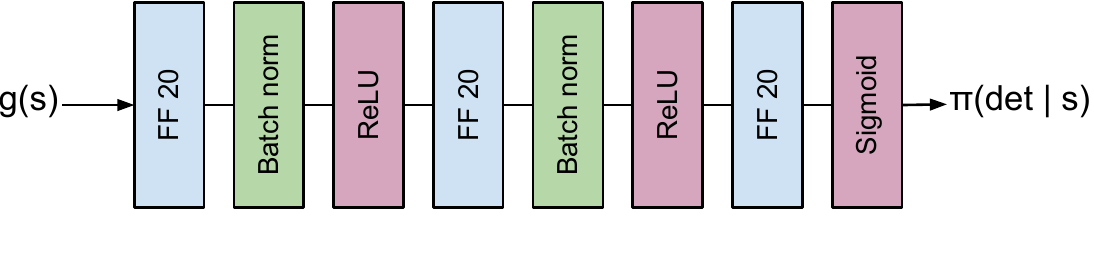}
    \caption{\textsc{pem} architecture of 3 feed-forward layers (20-hidden units each), ReLU activations and Batch Normalization. }
    \label{fig:pem-diag}
\end{figure}

We next learn a \textsc{pem} to act as a surrogate $\hat{f}_{\hat{\theta}}$ for our \textsc{yolo} perception system. Figure (\ref{fig:pem-diag}) shows our feed-forward \textsc{pem} architecture. To train, we first pre-process the data from our \textsc{yolo-kitti} detector, separating individual obstacles and pairing them to their closest bounding box prediction via Hungarian Matching \cite{forsyth2002computer}. For salient variables $g(s)$, we use 14 dimensions based on auxilliarly labels provided by the \textsc{kitti} dataset for each image. The 14 dimensions are: A 6-dimensional one-hot encoding of vehicle category, a 3-dimensional one-hot encoding for the amount the vehicle is occluded (None, Partial, Mostly), the $xyz$ obstacle location, and the rotation of the obstacle on the y-axis. After removing obstacles with unknown values, we were left with 37498 obstacles to train over. We then minimize loss function (\ref{eqn:pem-bce}) using the Adam Optimizer \cite{kingma2014adam} over $300$ epochs.

Note, our \textsc{pem} training has two limits. First, by considering labels on single obstacles, our model is unable to capture false positive detections. This is a common limitation \cite{sadeghi2021step}, but could be overcome with a more expressive surrogate. Second, we assume the \textsc{kitti} auxiliary labels are expressive enough to explain variance in detection performance across the domain. This is likely untrue---perception systems are affected by additional factors such as time of day, weather conditions, road surface reflections etc. In an industry-scale system deployment, we would expect to have dedicated teams of annotators to design and record a more widely expressive set of salient variables for each image. These issues lie outside the scope of this paper.

Table (\ref{tab:obstacle-BCE}) summarizes our \textsc{pem}'s performance (ML-NN), via 5-fold cross-validation. Several baselines are also given: LR fits a logistic regressor; Guess-$\mu$ returns the mean detection probability. We also fit a complex Bayesian Neural Network (B-NN). Rather than finding the most likely $\hat{\theta}$, B-NN learns a distribution over parameters using variational inference \cite{jospin2022hands}. We report the binary cross-entropy (\textsc{bce}) and area under the receiver operating characteristic curve (\textsc{roc-auc}) for each model. Across baselines, our original model ML-NN performs best.

It is perhaps surprising that B-NN does not perform better. It may be that the variational family used to capture the detection distribution was not sufficiently expressive. Alternatively, held-out data used in the 5-fold validation may be insufficiently different from the training data to realise the benefit of marginalizing over parameters. While not the focus of this paper, it could be interesting to investigate the performance of such \textsc{pem}s in cases of distribution shift and out-of-domain images.

\begin{table}
    \caption{Results of 5-fold cross validation across perception error models for KITTI obstacle detection task.}
    \ra{1.1}
    \centering
    \begin{tabular}{@{}lll@{}}
        \toprule
         Model & \textsc{bce} & \textsc{roc-auc}\\
        \midrule
         B-NN & 0.238 & 0.726\\
         LR &  0.244 &  0.688\\
         ML-NN & \textbf{0.226} & \textbf{0.777}\\
         Guess-$\mu$ & 0.260 & 0.5\\
    \end{tabular}
    \label{tab:obstacle-BCE}
\end{table}


\subsection{Adaptive Sampling of Rare Crash Events}

We now use our \textsc{pem} to drive adaptive sampling in algorithm (\ref{alg:full-system}). We run $K=10$ stages, with $N_{\kappa} = 100$, and $N_e = 100$. Each simulation runs for $T=100$ time steps, with a delta $0.05$. Cars start at 70 km/h, 15 metres apart.

Proposal network $q_{\phi}$ is a two-layer feed-forward network, with $32$ hidden nodes per layer, and $ReLU$ activations. Transformation function $h(s)$ projects $s$ down to a single dimension---the distance (in metres) between vehicles. To ensure our initial proposal produces trajectories similar to the original \textsc{pem}, we pre-train $\phi$ with $100$ simulations from $\hat{f}_{\hat{\theta}}$. In adaptation stages, we minimize (\ref{eqn:smooth-is}) with $\alpha=0.1$. We use the Adam optimizer \cite{kingma2014adam} for $500$ epochs.

We compare our method ($\textsc{adaptive-pem}_{100}$) to several baselines: $\textsc{mc}_{100}$ and $\textsc{mc}_{10,000}$ estimate crash probability using 100 and 10000 raw monte-carlo samples respectively. We also compare against a naive custom importance sampler ($\textsc{naive-50}_{100}$ and $\textsc{naive-50}_{1000}$) which ignores state information, instead detecting obstacles with a flat $50\%$ probability. This is likely to produce more frequent crashes, but the trajectories may be unlikely under the target.

For our robustness metrics $r(\tau, \varphi)$ used to evaluate thresholds $\gamma{\kappa}$, we consider three variants---Classical Spatial Robustness ($r_{c}$), \textsc{agm} Robustness ($r_{a}$), and Smooth-Cumulative Robustness ($r_{s}$), with $\gamma = 0$ across all three.

\subsection{Results}

\begin{table}
    \caption{Estimated failure probabilities under different sampling strategies. Averaged over 10 repetitions, with standard errors given in brackets.}
    \ra{1.1}
    \centering
    \begin{tabular}{@{}llll@{}}
        \toprule
        Method & $\hat{\mu}$ & Failures & \textsc{nll} \\
        \midrule
         $\textsc{mc}_{100}$ & 0.0 & 0 / 100 & -\\
         $\textsc{mc}_{10000}$ & 0.0 & 0 / 10000 & -  \\
         $\textsc{naive-50}_{100}$ & 2.64e-48 \begin{scriptsize} ($\pm$ 4.69e-49) \end{scriptsize}  & 77 / 100 & 199.95 \\
         $\textsc{naive-50}_{1000}$ & 4.75e-44 \begin{scriptsize} ($\pm$ 2.16e-44) \end{scriptsize}  & 748 / 1000 & 177.97 \\
         $\textsc{adaptive-pem}_{100}$ & 5.02e-15 \begin{scriptsize} ($\pm$ 3.83e-15) \end{scriptsize} & $52 / 100$ & $\mathbf{41.6}$
    \end{tabular}
    \label{tab:proposal-results}
\end{table}

Table \ref{tab:proposal-results} shows the results of each sampling strategy. Clearly, raw monte-carlo sampling is insufficient for this problem --- After 10,000 simulations, $\textsc{mc}_{10000}$ has produced 0 failures, reporting $0\%$ crash probability. This is unsurprising given the theoretical results around relative-error combined with the high accuracy of our target detector ($> 90\%$ average detection accuracy). Since conducting enough monte-carlo simulations to extract a ground-truth probability is computationally infeasible, we instead rely on two proxies---Proportion of final trajectories resulting in a crash, and average negative log likelihood (\textsc{nll}) of failures.

$\textsc{adaptive-pem}_{100}$, successfully learns a proposal that produces more than half of its simulations as failures, with an average $\textsc{NLL}$ of 41.6. In contrast, while $\textsc{naive-50}$ produces a larger failure proportion, the $\textsc{NLL}$ is far higher ($199.95$). This means trajectories from $\textsc{naive-50}$ are $\num{1e69}$ times less likely under $\hat{f}_{\hat{\theta}}$ than those from $\textsc{adaptive-pem}$.  As a result, \textsc{naive-50} is dominated by a few trajectories with large importance weights, leading to inaccurate estimates of $\hat{\mu}$. In fact, we can see that as the number of simulations increases from $100$ to $1000$, $\textsc{naive-50}$'s estimate gradually moves towards $\textsc{adaptive-pem}$'s estimate.

\begin{figure}
    \centering
    \includegraphics[width=0.6\linewidth]{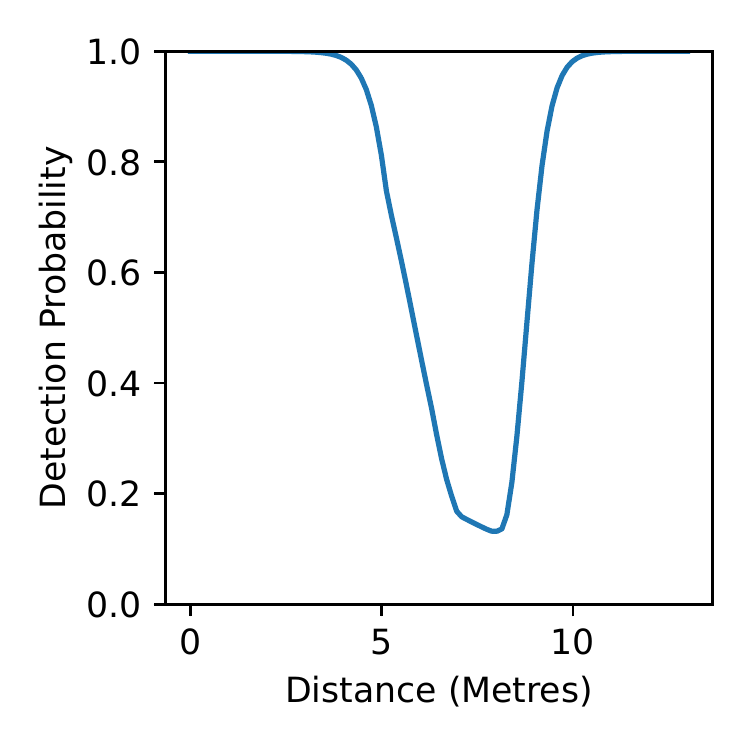}
    \caption{Learned proposal at $\kappa=10$ using Alg (\ref{alg:full-system})}
    \label{fig:proposal-dist}
\end{figure}

Figure (\ref{fig:proposal-dist}) shows the detection rate of \textsc{adaptive-pem} as a function of vehicle distance. At distances where safety is not a concern, the proposal mimics the detection accuracy of the original \textsc{pem} (between $98-99.9\%$). As the cars approach distances where a failure to brake would cause a crash, the proposal gradually lowers the detection probability.

\begin{table}
    \caption{\textsc{stl} metrics comparison}
    \ra{1.1}
    \centering
    \begin{tabular}{@{}llll@{}}
        \toprule
        Metric & $\hat{\mu_{\textsc{is}}}$ & Failures & NLL \\
        \midrule
         $r_{c}$ & 5.02e-15 \begin{scriptsize}($\pm$ 3.83e-15) \end{scriptsize} & $52 / 100$ & $41.6$\\
         $r_{a}$ & 6.76e-15 \begin{scriptsize} ($\pm$ 3.64e-15) \end{scriptsize} & $41 / 100$  &  $42.4$\\
         $r_{s}$ & 2.01e-15 \begin{scriptsize} ($\pm$ 5.38e-16) \end{scriptsize} & $40 / 100$  & $41.3$\\
    \end{tabular}
    \label{tab:STL-results}
\end{table}

\begin{figure}
    \centering
    \includegraphics[width=0.8\linewidth]{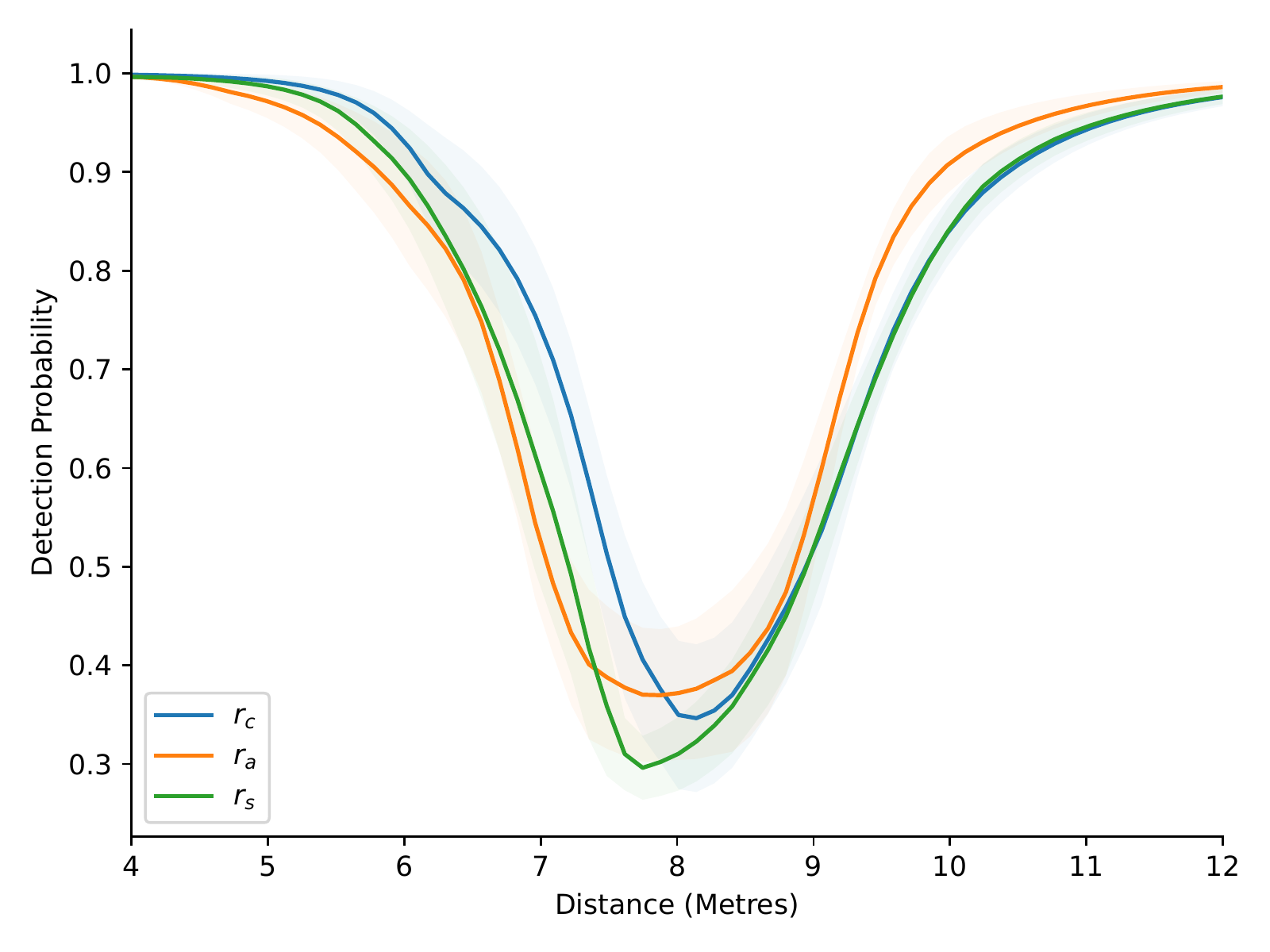}
    \caption{Proposal distributions learned for $r_c$, $r_a$, and $r_s$ at intermediate adaptation stage $\kappa=3$. Averaged over 10 repetitions.}
    \label{fig:stl-curves}
\end{figure}

\begin{figure}
    \centering
    \includegraphics[width=0.8\linewidth]{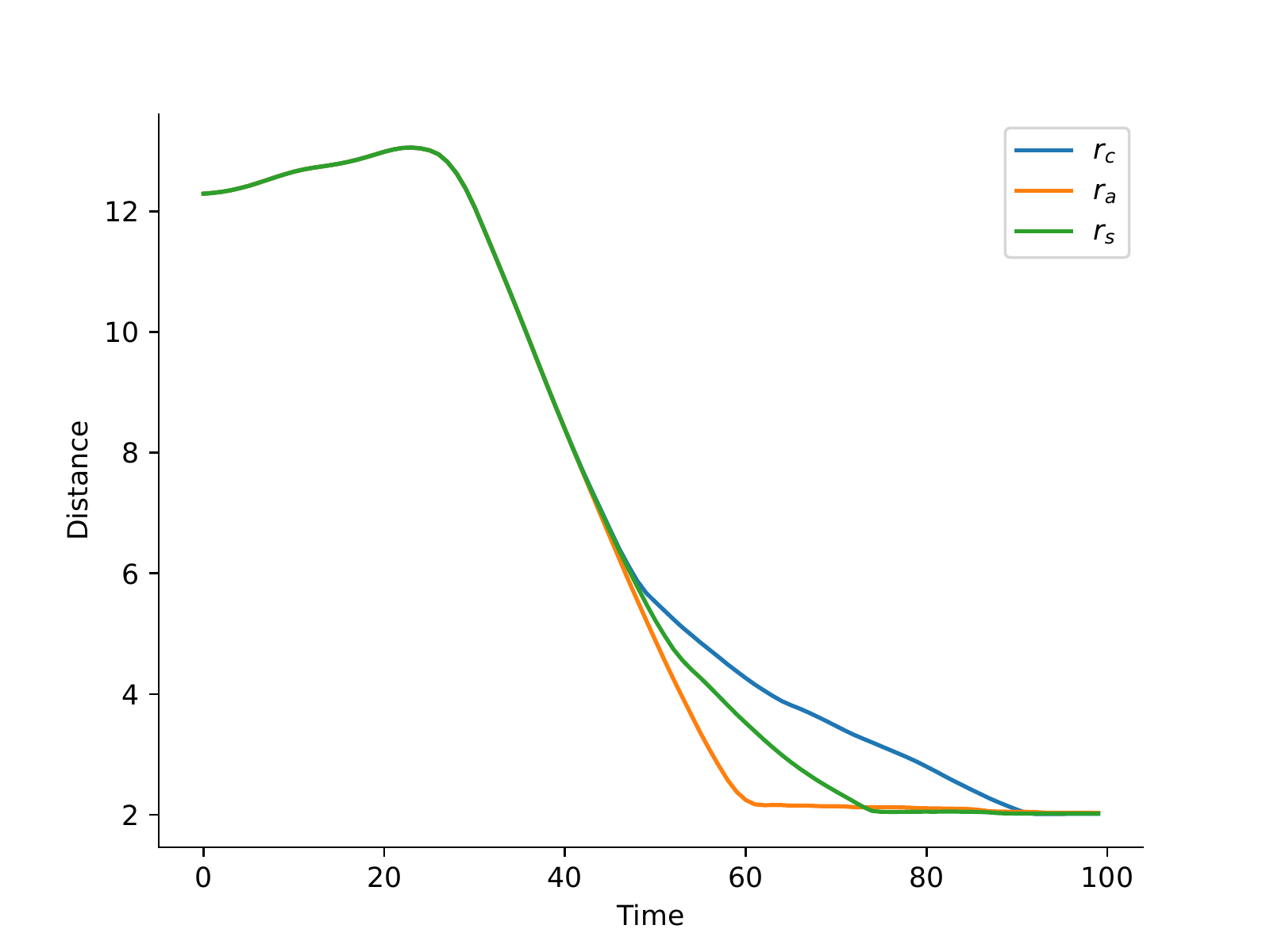}
    \caption{Trajectories for the three lowest ranked simulation runs according to the $r_c$, $r_a$, and $r_s$ robustness metrics.}
    \label{fig:rollouts-comparison}
\end{figure}

For robustness metrics $r(\tau, \varphi)$, Table (\ref{tab:STL-results}) shows that the final proposals and $\hat{\mu}$-estimates after $K=10$ adaptation stages is roughly similar across all three. However, as figure (\ref{fig:stl-curves}) reveals, the behaviour of the three metrics at intermediate stages can vary.

To understand why this occurs, consider the different ways $r_c$, $r_a$, and $r_s$ rank the safety of a batch of simulations. Figure (\ref{fig:rollouts-comparison}) shows the trajectories that each metric chose as ``least safe'' from a collection of 10000 random simulations. These choices align with our understanding of how each metric calculates robustness: $r_c$ computes robustness as a strict minimum of distances across all time steps, so chooses the trajectory with the single smallest distance value. In contrast, $r_a$ applies a geometric mean over distances, leading the metric to choose a trajectory with a (marginally) higher minimum distance, but which stays lower for a longer time-span. Finally, $r_s$ applies a smoothed approximation to the minimum, leading it to choose a trajectory somewhere between the absolute minimum and mean values. These ranking differences affect trajectory sorting in line 6 of Algorithm (\ref{alg:full-system}), which then influences the trajectories chosen in each adaptation stage to optimize (\ref{eqn:smooth-is})

The main insight is, given limited adaptation budget, the choice of $r(\tau, \varphi)$ can influence threshold levels $\gamma_{\kappa}$, trajectory ordering, and potentially learning process behaviour.

\section{RELATED WORK}

Many works test systems via black-box falsification or adaptive-stress testing \cite{dreossi2018semantic, dreossi2019compositional, annpureddy2011s, julian2020validation, lee2020adaptive, Delecki2022HowDW}. Such works focus on finding a single failure trajectory, rather than estimating the overall probability of failure. 

This paper estimated rare-event failures by augmenting \textsc{cem} techniques \cite{kroese2013cross}. Other works use \textsc{cem} for AV risk estimation \cite{zhao2017accelerated, o2018scalable, kim2016improving}. These typically learn proposals over trajectories, an approach which can fail when rare events result from a product of probabilities over long-running trajectories. Additionally, these works often limit themselves to exponential-family proposals over a set of initial conditions; They avoid sampling perceptual behaviours of the vehicle itself. Our method instead learns proposals over individual states, using \textsc{pem}s to handle sensor observations.

One state-dependent approach close to ours frames rare-event sampling as an approximate dynamic programming problem \cite{corso2020scalable}. It requires the user to have prior knowledge of the transition model, and the ability to jump to arbitrary states. For black-box simulators like CARLA, neither requirement holds.

An alternative adaptive sampling technique is multi-level splitting \cite{norden2019efficient, sinha2020neural}. Here initial simulation trajectories are treated as particles which are perturbed using Markov-Chain Monte Carlo (\textsc{mcmc}). This method scales better to long-running sequential problems, but conducting \textsc{mcmc} transitions per trajectory per stage is intractable for simulators with non-trivial run time.

\emph{Continuation approaches} use earlier perception system training epochs as a proxy for proposal stages \cite{uesato2018rigorous}. They assume the user has access to earlier training checkpoints, and that failures in those earlier stages are similar to those encountered by the final system.

\textsc{pems} are applied in \textsc{av} as surrogates for image, radar, and lidar sensors \cite{sadeghi2021step, mitra2018towards, piazzoni2020modeling, berkhahn2021traffic}. The primary motivation of such applications is usually to bypass expensive rendering, providing fast, low-fidelity predictions. In contrast, our work takes advantage of the probabilistic outputs provided by \textsc{pem}s to also drive adaptive importance sampling. 

We make the core assumption that our \textsc{pem} is \emph{well calibrated} \cite{gawlikowski2021survey}. Several issues can challenge this assumption. First, our model may encounter \emph{epistemic uncertainty} if images in simulation are far from the training distribution \cite{sensoy2018evidential}. Second, our model may encounter \emph{aleatoric uncertainty} if the salient variables insufficiently express relevant safety aspects of a scene \cite{kendall2017uncertainties}. In future work, we aim to explore this relationship between epistemic/aleatoric uncertainty in the upstream perceptual system, and the guarantees given on downstream failures.

There have been multiple attempts to encode complex traffic rules using temporal logic \cite{gressenbuch2021predictive, maierhofer2022formalization, hekmatnejad2019encoding, arechiga2019specifying}. In future work, we aim to scale our method from automated braking to more complex traffic scenarios, investigating the effect of balancing multiple competing rules.

For testing of black-box systems, one issue is assessing trust in your final estimation. Safety certificates help ensure risk is not under or over-estimated \cite{fan2020statistical, younes2005probabilistic, arief2021deep, harper2021safety}. These techniques are complementary to our task, and can be applied as an additional step after sampling.

\section{CONCLUSION}

This paper proposed a method for testing black box perceptual-control systems in simulation, even when the perception system takes real-world sensor inputs, and failures are rare. We used \textsc{pem}s to drive state-dependent adaptive importance samplers, with an \textsc{stl} robustness metric to assess failure thresholds. Our experiments in automated braking showed our method efficiently estimates rare failure probabilities, learning distributions that generate a high percentage of high-likelihood failure trajectories.


\bibliographystyle{IEEEtran}
\bibliography{IEEEabrv,references}
\end{document}